\documentclass{edm_article}

\usepackage{graphicx}
\usepackage{float}
\usepackage[caption = false]{subfig}

\DeclareMathOperator*{\argminA}{arg\,min} 

\usepackage{hyperref} 
\usepackage{breqn}
\usepackage{amsmath}

\begin{document}

\title{Fair Classification via Transformer Neural Networks:\\
Case Study of an Educational Domain}

\numberofauthors{2}
\author{
\alignauthor
Modar Sulaiman\\\
       \affaddr{University of Tartu}\\
       \affaddr{Tartu, Estonia}\\
       \email{modar.sulaiman@ut.ee}\\
\alignauthor
Kallol Roy\\
       \affaddr{University of Tartu}\\
       \affaddr{Tartu, Estonia}\\
       \email{kallol.roy@ut.ee}
}

\maketitle

\begin{abstract}
Educational technologies nowadays increasingly use data and Machine Learning (ML) models. This gives the students, instructors, and administrators support and insights for the optimum policy. However, it is well acknowledged that ML models are subject to bias, which raises concerns about the fairness, bias, and discrimination of using these automated ML algorithms in education and its unintended and unforeseen negative consequences. The contribution of bias during the decision-making comes from datasets used for training ML models and the model architecture. This paper presents a preliminary investigation of the fairness of transformer neural networks on the two tabular datasets: Law School and Student-Mathematics. In contrast to classical ML models, the transformer-based models transform these tabular datasets into a richer representation while solving the classification task. We use different fairness metrics for evaluation and check the trade-off between fairness and accuracy of the transformer-based models over the tabular datasets. Empirically, our approach shows impressive results regarding the trade-off between fairness and performance on the Law School dataset.
\end{abstract}

\keywords{Fairness; Classification; Educational Data; Bias; Representation Learning; Machine Learning; Transformer}

\section{Introduction}
Automated decision-making with ML models in education is increasingly used to aid and support teachers, educators, and other stakeholders for optimal policy formulation. Though this method holds immense potential to improve prediction accuracy, the outcomes of the ML models show unfair results to some groups (e.g., an underrepresented minority group) of society. For example, the ML models show unfair approval for student loans for African students or predict lower bar exam success of students of low socio-economic group \cite{fei2015temporal}. 

In general, the notion of unfairness in ML is broadly categorized as follows \cite{zafar2019fairness}, (i) Disparate treatment: where the ML model classifies differently (unfairly)  people with the same values of non-sensitive features but different values of sensitive features, (ii) Disparate impact: where the ML model classifies that benefits (or hurts) people who are sharing the value of a sensitive feature vector more frequently than the other group, (iii) Disparate mistreatment: where the ML model achieves different classification accuracy for groups of people sharing different values of a sensitive feature.\newline 
Our paper mainly investigated the fairness of classification using transformer models on tabular data in the educational domain. We used tabular non-textual data to train the transformer networks and thus verify their ability to make fair predictions in the classification tasks in two scenarios: first, training transformer networks without any bias mitigation method, and second, considering a bias mitigation method (Section \ref{fairness_constraints}) while training transformer networks. The representation of non-textual data in a rich representation via the transformer model is one of the main strengths of the transformer models. It follows the distributional hypothesis: a word is characterized by the company it keeps.

In our work, we showed that there exists a transformer model, namely the SAINT model, which achieves perfect group fairness without requiring any explicit debiasing method. In addition, one of the transformer models, the Tab model (with fairness constraint), improves fairness for the protected group at a negligible cost in terms of accuracy compared to the other models on the Law School dataset. Furthermore, we indicated a case using the Student-Mathematics dataset, where we do not recommend using the transformer models to mitigate bias in the final predictions. Finally, we demonstrated the possibility of empirically achieving a slight trade-off between performance and fairness using transformer models.

To the best of our knowledge, our work is the first study that explores the fairness ability of transformer-based models on tabular data in educational and other domains—in addition to considering the use of fairness constraint to train transformer-based models on tabular data in particular.

\section{RELATED WORK}
A comprehensive studies of algorithmic fairness in education is reported in \cite{DBLP:journals/corr/abs-2007-05443}. They have investigated how discrimination emerges in automated systems and how it can be mitigated through studying and calibrating: the measurement of input data, model learning, and output presentation. A comparative study of different fairness algorithms on multiple datasets are reported  \cite{https://doi.org/10.48550/arxiv.1802.04422}). \newline
The Law School Admission Council (LSAC) National Longitudinal Study reported discrimination against Africans (and other minorities) examinees during the bar passage examinations \cite{wightman1998lsac}. Recent approaches of using transformer-based models are showcased in \cite{gorishniy2021revisiting,huang2020tabtransformer, somepalli2021saint} and in Section \ref{all_models} we provide details of those transformer-based models. However, in contrast to the previous works \cite{gorishniy2021revisiting,huang2020tabtransformer, somepalli2021saint} which only considered the performance of transformer-based models, we validate the fairness capability and performance of transformer-based models simultaneously. A flurry of interesting and insightful research on fairness constraints are investigated recently by various research groups \cite{zafar2019fairness,liu2020accuracy,agarwal2018reductions}. Several authors addressed the problem of trade-off between fairness and accuracy \cite{rodolfa2021empirical, dutta2020there} by explaining why a trade-off exists on a given biased dataset and demonstrating the possibility of empirically improving fairness without sacrificing accuracy.
\section{FAIR CLASSIFICATION THEORY}\label{fairness_constraints}
Fairness of machine learning examines the disparate treatment, disparate impact, or disparate mistreatment of different groups in a population, which are devised based on sensitive attributes, denoted by $z$, such as race, etc. Groups are divided into privileged groups (favorable treatment) and unprivileged groups (unfavorable treatment). Without loss of generality, let $z = 0$ be the unprivileged group and $z = 1$ be the privileged group. Privileged population groups would be more likely to receive favorable treatment than unprivileged population groups by a biased model. \\
In general, the binary classification task can be formulated as finding the optimum mapping function $\mathcal{T}$ using input features $x \in \mathcal{R}^{d}$ and class labels $y \in (0,1)$. For example, the label $y = 1$ could represents the application loan is getting accepted and $y = 0$ otherwise. This decision boundary-based classifier $\mathcal{T}$ predicts a label $y$, by minimizing the cross-entropy loss function over a training set:
\begin{equation}\label{eq:1.1}
\argminA_{w}\mathcal{L}_{w}W
\end{equation}
\begin{equation}\label{equ1}
\mathcal{L}(\mathcal{T}(x)=\hat{y}, y) = -{(y\log(\hat{y}) + (1 - y)\log(1 - \hat{y}))}
\end{equation} 
where $\hat{y}$ is the predicted label and $w$ is the set of model weights. However, the classifier can be trained fairly using additional constraint for fairness guarantees of unprivileged group \cite{agarwal2018reductions, zafar2019fairness}. The fairness constrained optimization problem can be expressed as the following:
\begin{equation}
\begin{aligned}\label{eq:2}
\argminA_{w}\mathcal{L}_{w} \\
\textrm{subject to} \quad & P(.|z = 0) = P(.|z = 1) \\
\end{aligned}
\end{equation}
where the constraint in (\ref{eq:2}) can be formulated as a condition of \textit{no disparate treatment} (\ref{eq:3}), \textit{no disparate impact} (\ref{eq:4}) or \textit{no disparate mistreatment} (\ref{eq:5}, \ref{eq:6}, \ref{eq:7}) as shown below.
\begin{align} 
P(\hat{y}|x, z) = P(\hat{y}|x) \label{eq:3}\\ 
P(\hat{y} = 1|z = 0) = P(\hat{y}|z = 1) \label{eq:4} \\
P(\hat{y} \neq y|z = 0) = P(\hat{y} \neq y|z = 1) \label{eq:5}\\
P(\hat{y} \neq y|y = -1, z = 0) = P(\hat{y} \neq y|y = -1, z = 1) \label{eq:6}\\
P(\hat{y} \neq y|y = 1, z = 0) = P(\hat{y} \neq y|y = 1, z = 1)  \label{eq:7}
\end{align}  
We consider only the condition of \textit{no disparate treatment} (\ref{eq:3}) as the fairness constraint in our experiments. However, finding an optimum fair classifier defined by the constrained optimization problem (\ref{eq:2}) is non-trivial. The desired fair classifier (satisfying the constraints) may be of a non-convex boundary-based type; thus, finding the optimum weights $w$ in some cases is challenging.

\section{PROPOSED MODELS}\label{all_models}
This section highlights the models that we used in our work. Most of the transformer-based models are adaptations of the transformer architecture \cite{vaswani2017attention}, and we examine them for fair classification in the tabular datasets in the educational domain.\newline 
The first transformer-based model that is explicitly designed for tabular data is Tab-Transformer \cite{huang2020tabtransformer}. For the sake of brevity, we refer to the Tab-Transformer model as \textit{Tab} model in our paper. The Tab model uses attention mechanism to embed only categorical features in the tabular data. A very similar model to the Tab model is FT-Transformer (\textit{Feature Tokenizer + Transformer}) model \cite{gorishniy2021revisiting}, or \textit{FT} model, for short. The FT model is also designed specifically for tabular data. However, the FT model transforms all features (categorical and numerical) to embeddings where the continuous features are projected into a $d-$dimensional space before passing them through the transformer encoder.
\newline 
Our paper shows that the most critical transformer-based model regarding fairness in the tabular dataset is the \textit{SAINT} model, \textit{Self-Attention and Intersample Attention Transformer} \cite{somepalli2021saint}. The architecture of the SAINT model itself plays the role of a regularizer for achieving fairness. Each block of SAINT model consists of two attention blocks: self-attention transformer block \cite{vaswani2017attention} and intersample attention (a type of row attention) transformer block. The intersample attention block of SAINT model computes attention across samples. In this approach, the features of a given data point interact with each other, then data points interact with each other using entire rows. Consequently, the SAINT model implicitly satisfies the condition (\ref{eq:3}) since it provides a better contextual representation than the other models for the data points and thus, does not need to add a fairness constraint to the objective function in some cases. \newline 
The last transformer-based model we used is Perceiver \cite{jaegle2021perceiver}. Perceiver model is designed to be architecture agnostic of the nature of the input data. It handles arbitrary configurations of different modalities (audio, images, and text). In our experiment we check the performance and fairness of Perceiver model in tabular data modality. In addition to the previous transformer-based models, we used one of the classical ML models, Logistic Regression (LR) model. LR is one of the common statistical analysis methods to predict a binary outcome and used in different works when studying fairness of ML.

\section{Datasets and Fairness Metrics}\label{Datasets_and_Fairness_Metrics}
In this section, we describe the datasets used in our experiments and define the fairness metrics for assessing the models' fairness. 

\subsection{Datasets}\label{Datasets}
We used two tabular datasets in our experiments: Law School (LSAC) and Student-Mathematics datasets. For both datasets, we consider using the same features and processing\footnote{\url{https://github.com/tailequy/fairness-dataset/tree/main/experiments}} that are used to conduct the experiments in the survey \cite{le_quy_survey_2022}. \newline
The Law School dataset was developed by a Law School Admission Council (LSAC) survey across 163 law schools in the United States in 1991 \cite{wightman1998lsac}. The dataset is investigated in a variety of studies and is currently hosted in the database of Project SEAPHE\footnote{\url{http://www.seaphe.org/databases.php}}. After cleaning and processing the law school data, we got 20,798 samples of students. We used 12 attributes (3 categorical, 3 binary and 6 numerical attributes) for Law School dataset in experiments. The model is used to predict whether or not an examiner will pass the bar exam. We label 'Non-white' to be the minority and unprivileged group in our experiments, 16\% of the Law School dataset. \newline
In the Student-Mathematics dataset\footnote{\url{https://archive.ics.uci.edu/ml/datasets/student+performance}}, the task is to predict the performance of secondary school students in mathematics subject \cite{cortez2008using}. The dataset contains 395 samples of students with 33 features (4 categorical, 13 binary and 16
numerical attributes). The female students in the Student-Mathematics dataset are the majority group, with about 52.7\% of the dataset.

\subsection{Fairness Metrics}
Our paper focused on the disparate treatment of population groups (privileged and unprivileged), so we used different group fairness metrics to determine the bias in the model's final predictions. \newline
We investigated three group fairness metrics in total: Absolute Between-ROC Area ($ABROC$) \cite{gardner2019evaluating}, Equal Opportunity Difference ($EOD$) \cite{hardt2016equality} and Statistical Parity Difference ($SPD$) \cite{dwork2012fairness}. We interpret that the more unfair a model, the greater the difference in fairness metric values of each subgroup. \newline
The $ABROCA$ statistic is based on the Receiver Operating Characteristics (ROC) curve, and measures the absolute value of the area between the baseline group ROC curve $ROC_{b}$ and those of one or more comparison groups $ROC_{c}$. Whereas $EOD$ measures the difference between True Positive Rates (TPR) for unprivileged and privileged groups. Finally, $SPD$ measures the difference between the probability of unprivileged group gets favorable prediction and the probability of privileged group gets favorable prediction.

\section{EXPERIMENTS AND RESULTS}\label{EXPERIMENTS_RESULTS}
\begin{figure*}
\subfloat[{\scriptsize LR with\\
fairness constraint}]{\includegraphics[width = 1.5in]{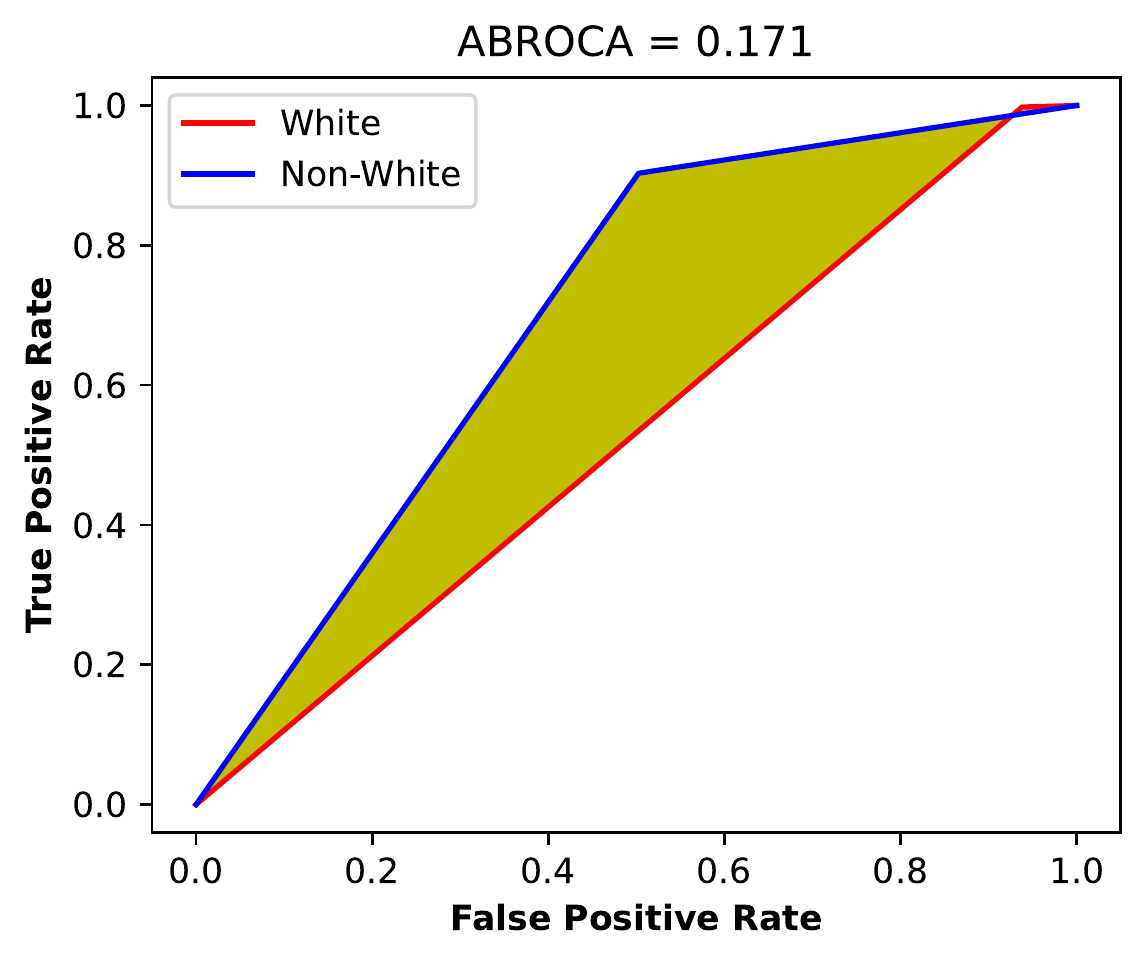}} 
\subfloat[{\scriptsize Tab-Transformer without\\
fairness constraint}]{\includegraphics[width = 1.5in]{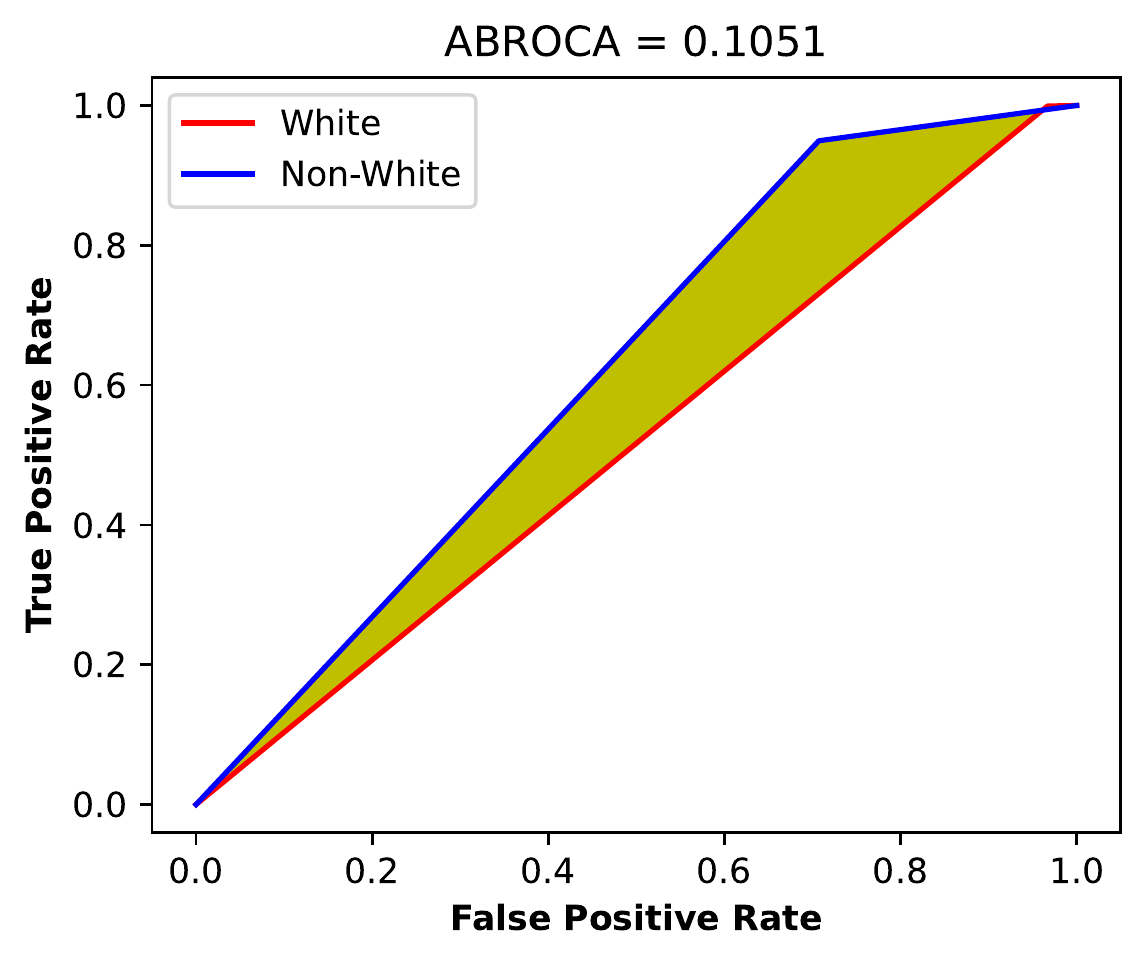}}
\subfloat[{\scriptsize FT-Transformer without\\
fairness constraint}]{\includegraphics[width = 1.5in]{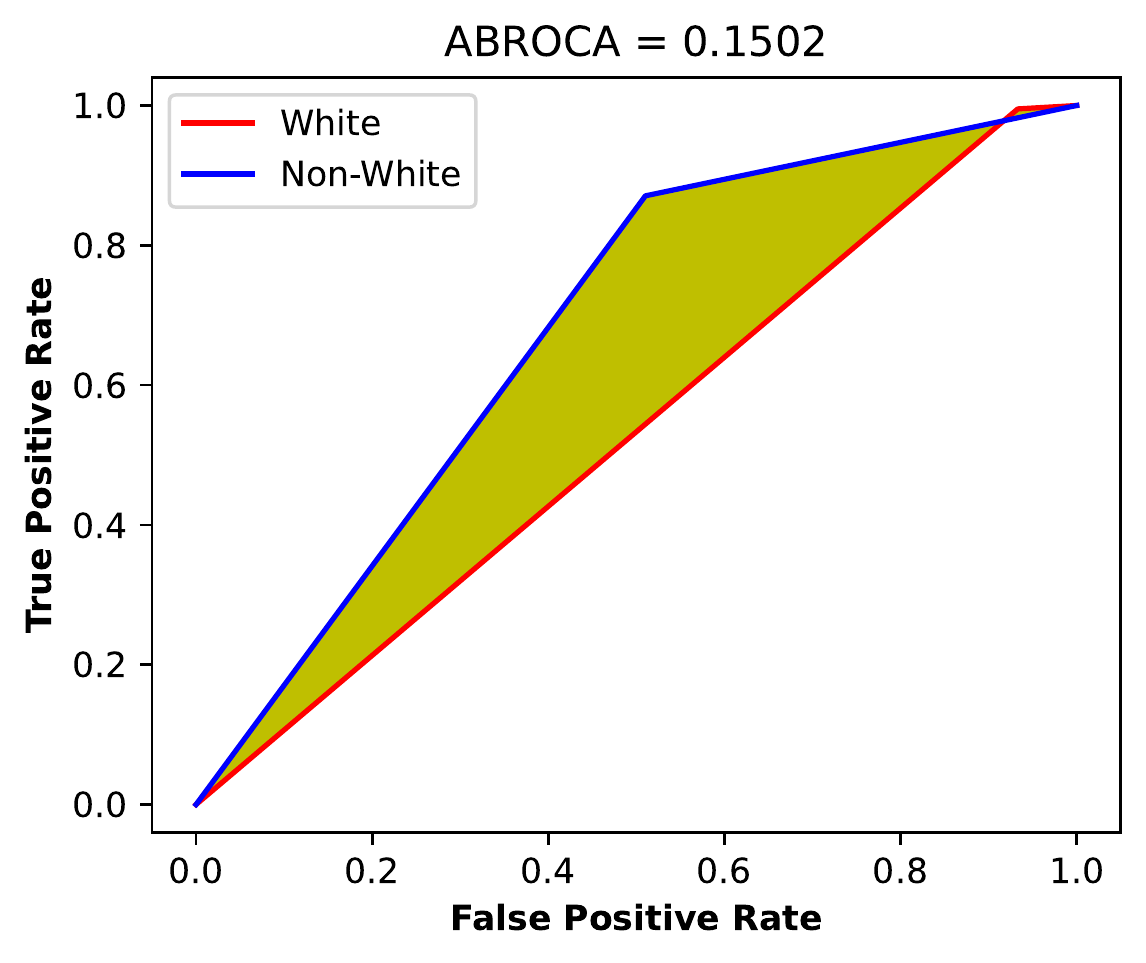}}
\subfloat[{\scriptsize SAINT without\\
fairness constraint}]{\includegraphics[width = 1.5in]{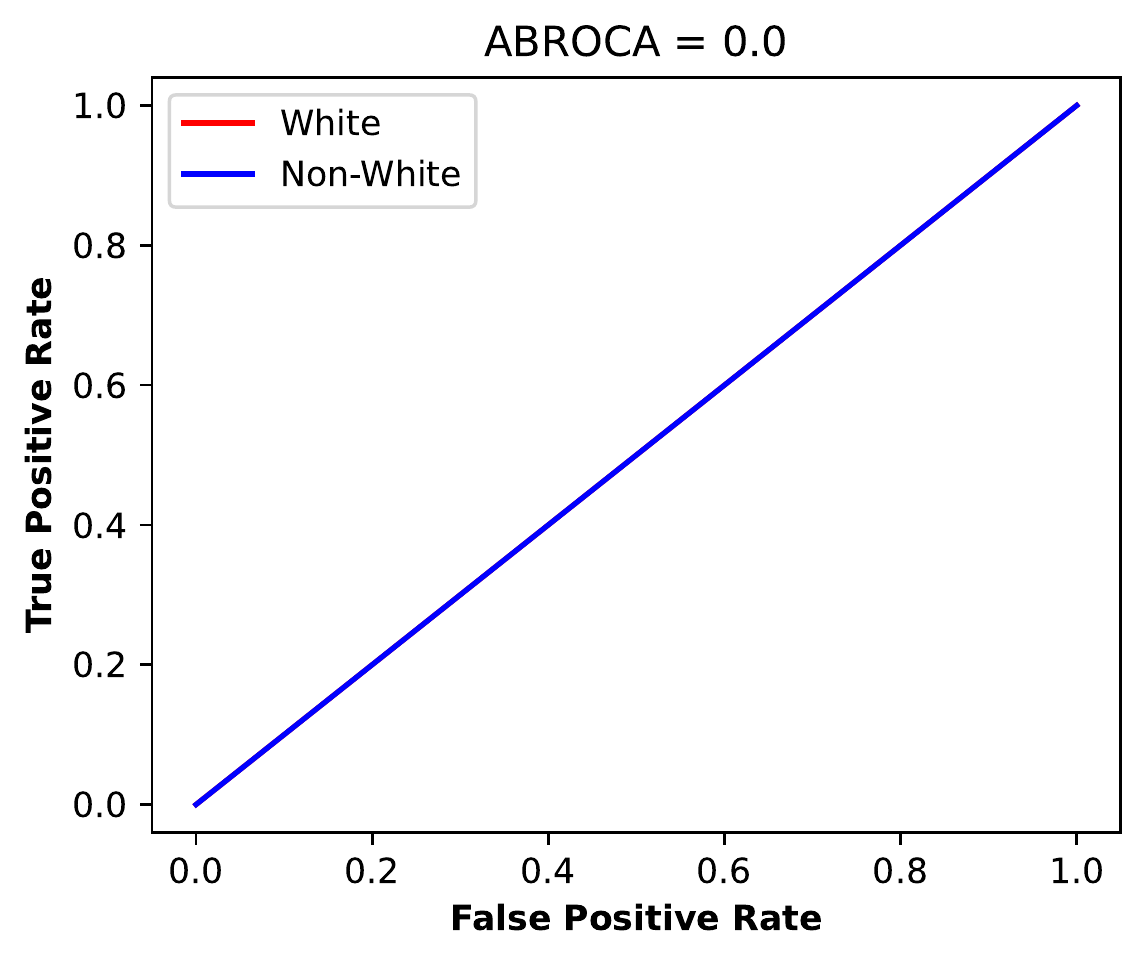}} 
\subfloat[{\scriptsize Perceiver without\\
fairness constraint}]{\includegraphics[width = 1.5in]{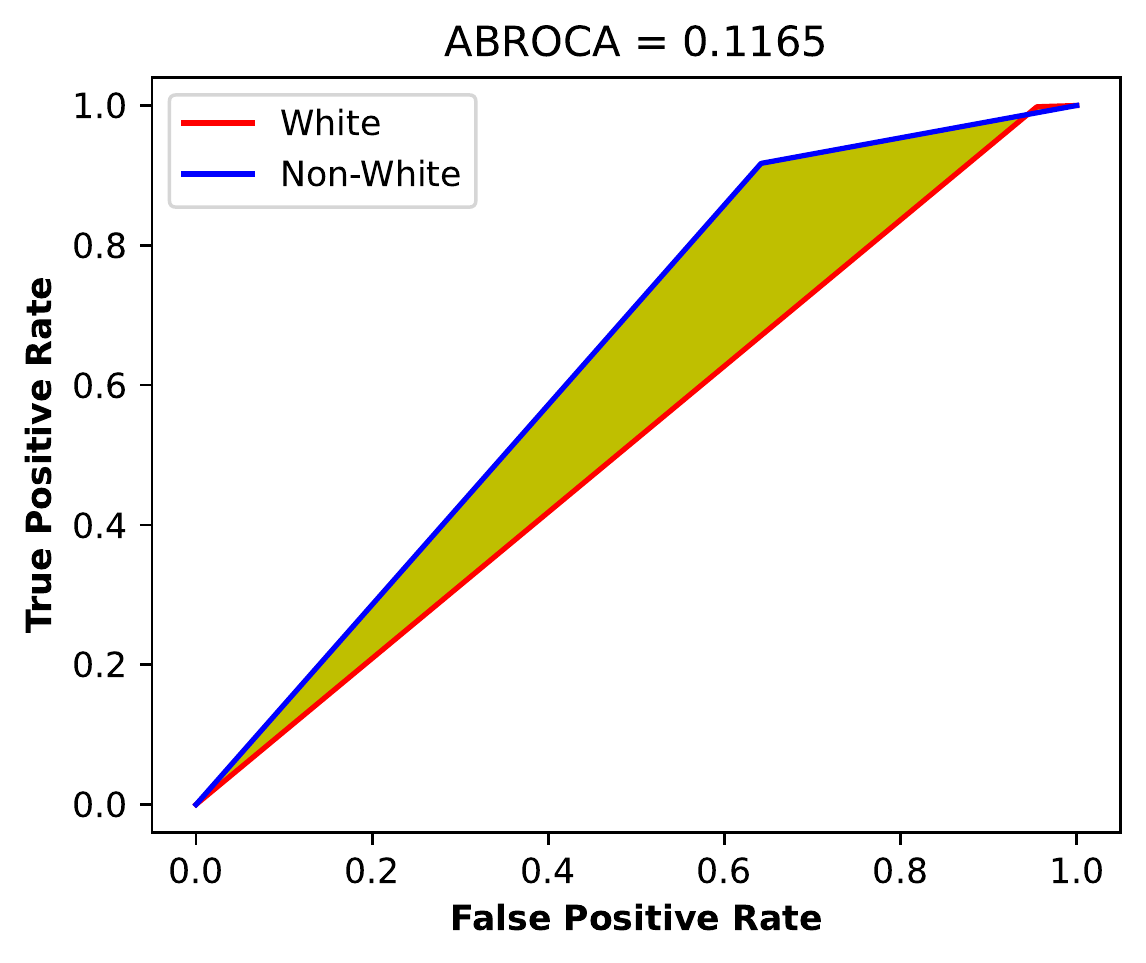}}\\
\subfloat[{\scriptsize LR with\\
fairness constraint}]{\includegraphics[width = 1.5in]{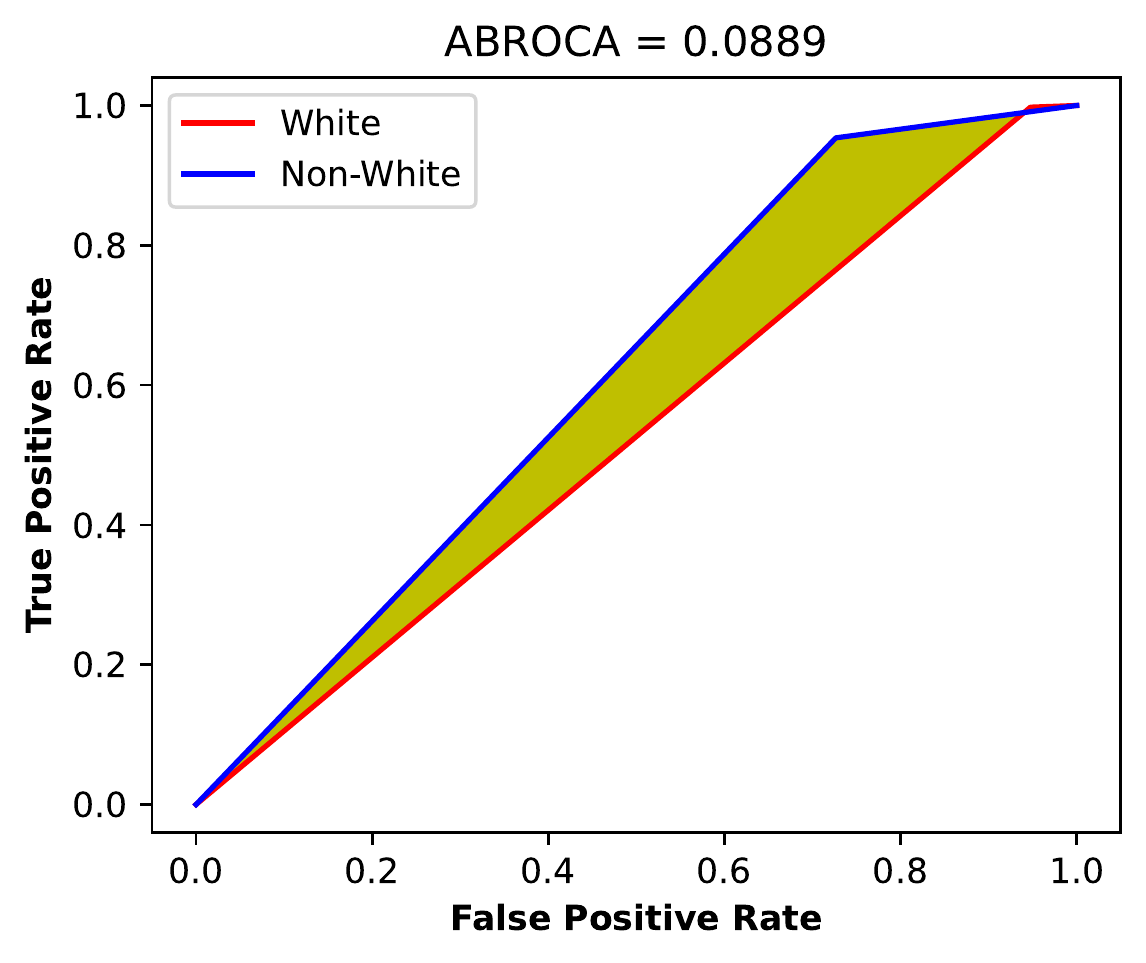}} 
\subfloat[{\scriptsize Tab-Transformer with\\
fairness constraint}]{\includegraphics[width = 1.5in]{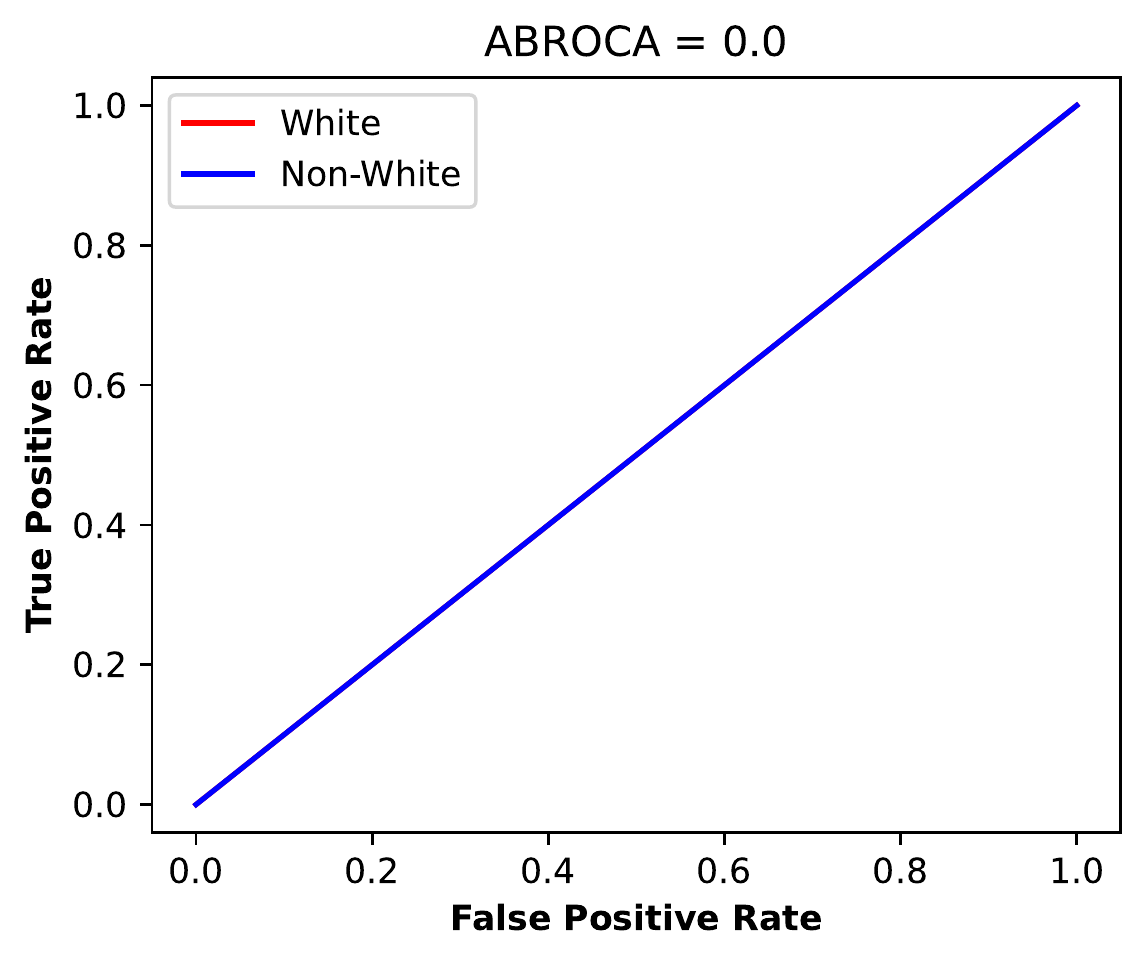}}
\subfloat[{\scriptsize FT-Transformer with \\
fairness constraint}]{\includegraphics[width = 1.5in]{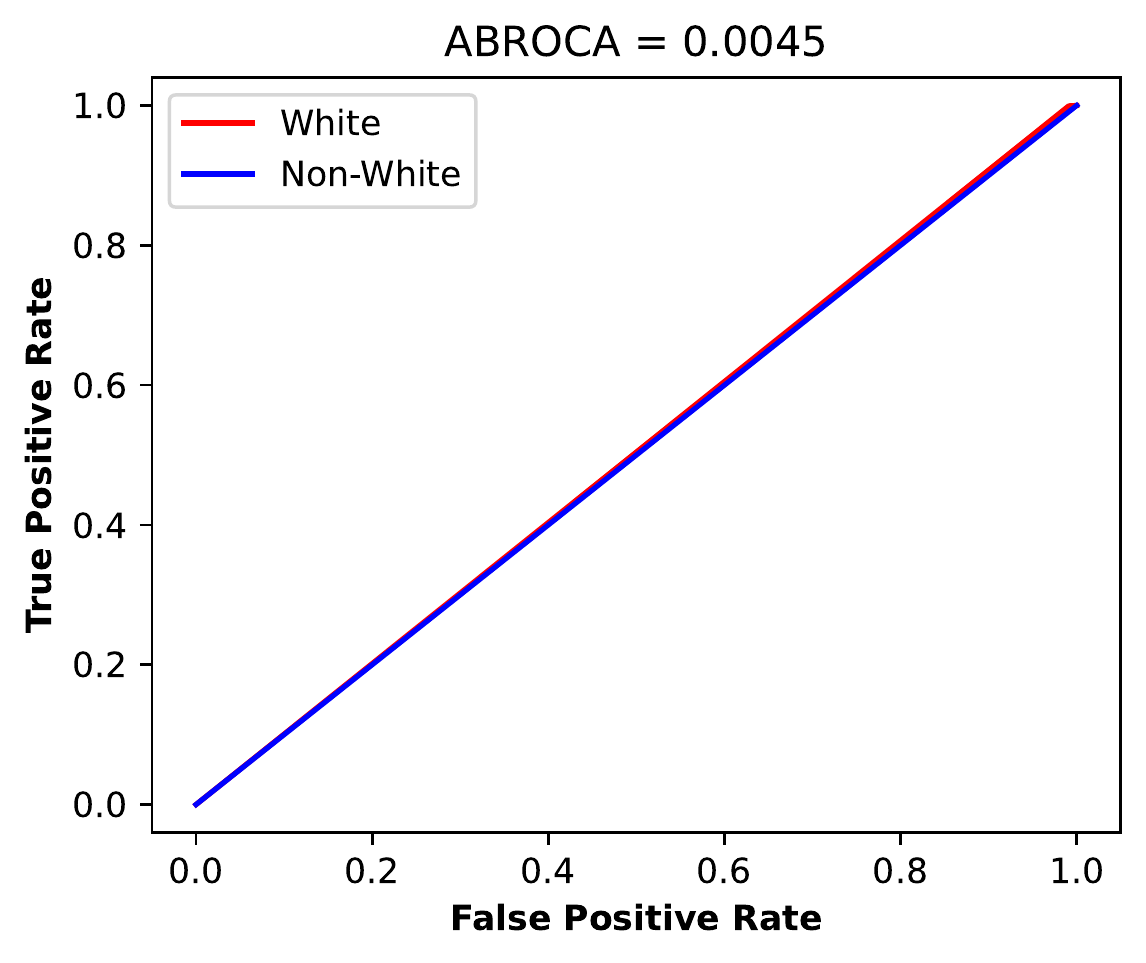}}
\subfloat[{\scriptsize SAINT with fairness \\
constraint}]{\includegraphics[width = 1.5in]{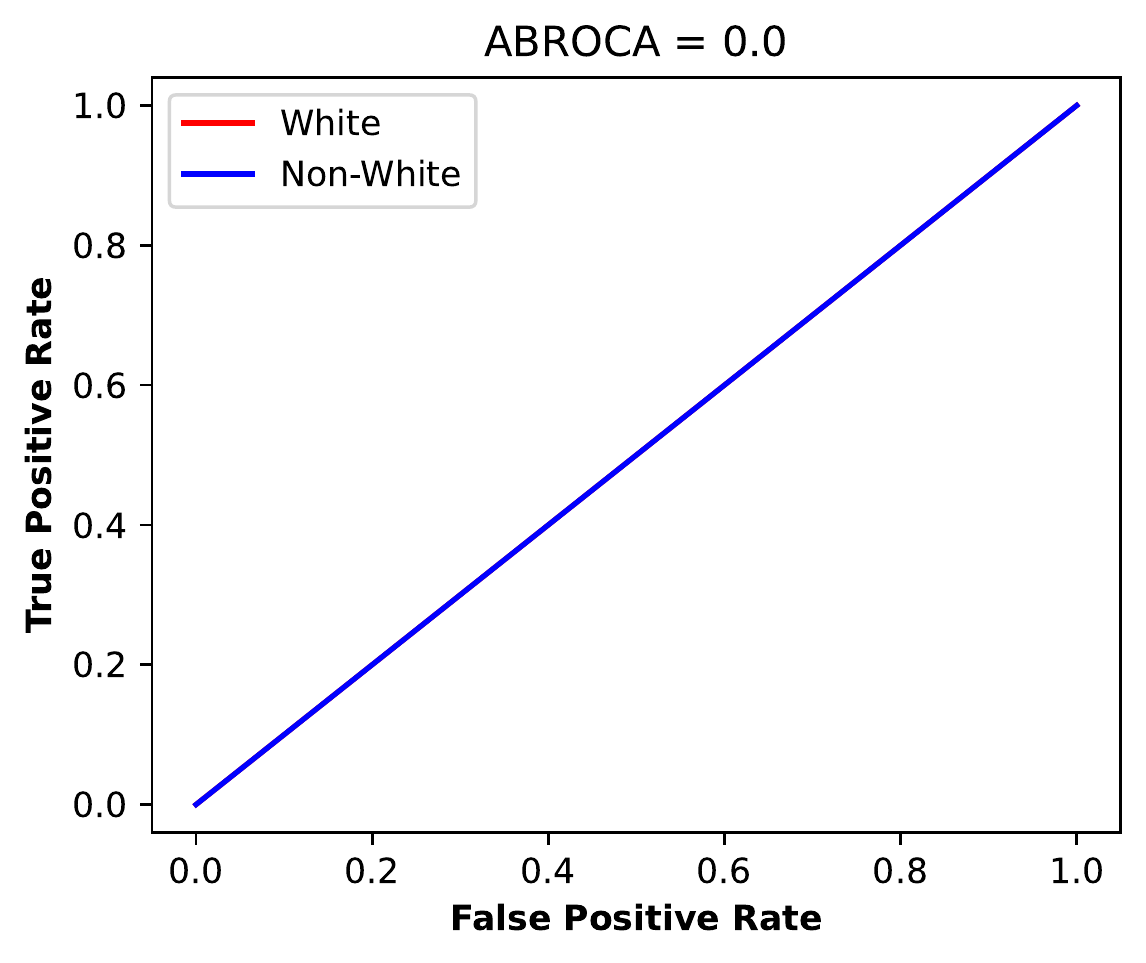}} 
\subfloat[{\scriptsize Perceiver with fairness\\
constraint}]{\includegraphics[width = 1.5in]{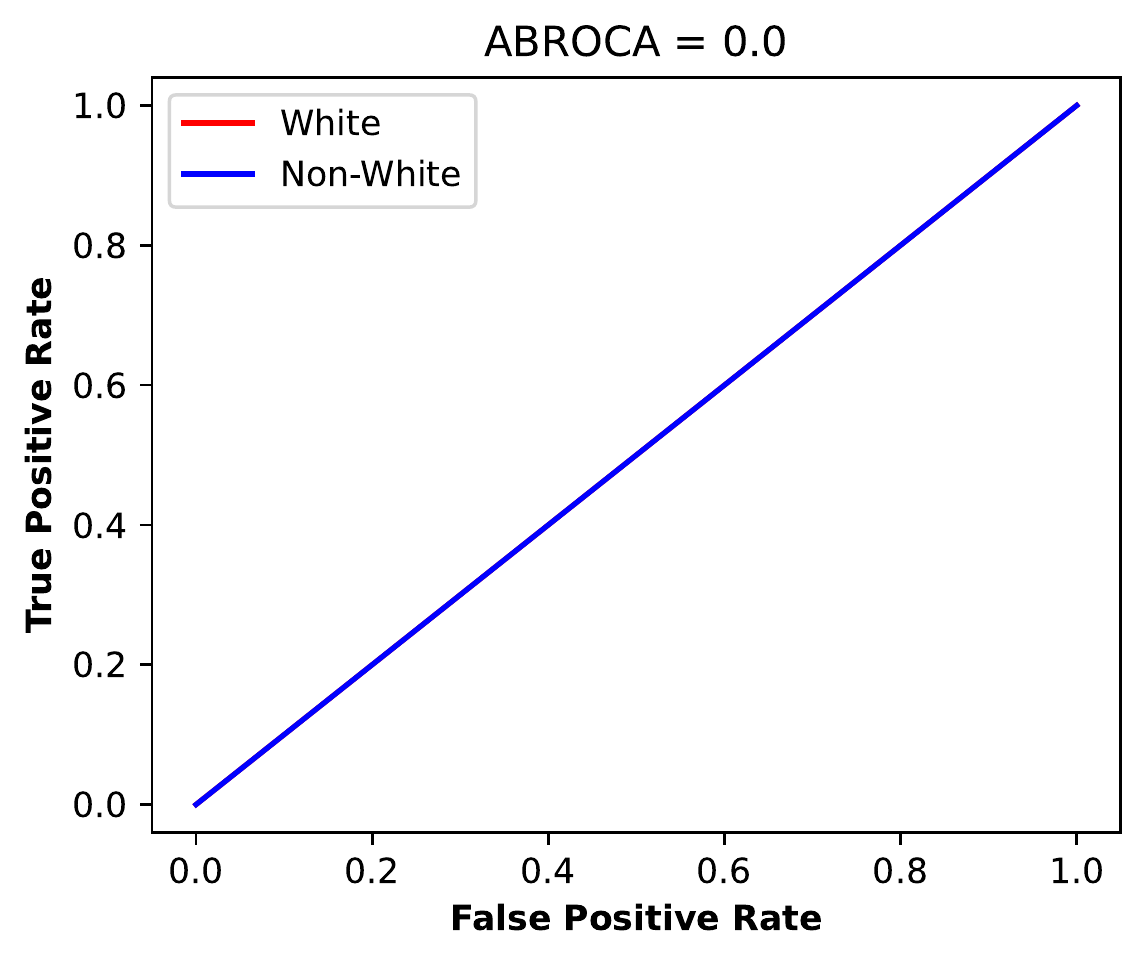}} 
\caption{ABROCA of each model on Law School Data. Figures (a, b, c, d, e) show ABROCA of each model without applying fairness constraints. Figures (f, g, h, i, j) show ABROCA of each model when applying fairness constraints.}
\label{fig:ABROC_models}
\end{figure*}

This section presents the results of our experiments and illustrates the trade-offs between fairness and performance. We randomly split the Law school and Student-Mathematics datasets into training and test sets for our experiments. Each test set was chosen to have 30\% of the samples and the training sets to contain the remaining samples from each dataset. \newline 
\begin{table}[h!]
\begin{tabular}{ |p{1.2cm}|p{1.1cm}|p{1.5cm}|p{1.4cm}|p{1.4cm}|}
\hline
\hline
\multicolumn{5}{c}{Dataset: Law School. Protected attribute: race.} \\
\hline
\hline
Model &  F1  &  Accuracy & SPD &   EOD\\
\hline
\hline
LR & \textbf{0.94984}    &\textbf{0.90721} &  0.189538   &  0.082670  \\\hline
FT  &  0.94504   & 0.89839 &  -0.215906 & -0.124452 \\
\hline
Tab   &  0.94664  & 0.90016 &  -0.112048 &-0.049809\\

\hline
Perceiver   & 0.94590   & 0.89919 & -0.151387  & -0.081128 \\
\hline
SAINT   & 0.94299 & 0.89214 & \textbf{0} & \textbf{0}\\
\hline
\hline
\end{tabular}
\begin{tabular}{ |p{1.2cm}|p{1.1cm}|p{1.5cm}|p{1.4cm}|p{1.4cm}|}
\multicolumn{5}{c}{Dataset: Student-Mathematics. Protected attribute: sex.} \\
\hline
\hline
Model &  F1  & Accuracy & SPD &   EOD\\
\hline
LR & \textbf{0.91111} &\textbf{0.93277} &  0.153193  & -0.005847 \\\hline
SAINT   & 0.76041 & 0.61344 &  \textbf{0} & \textbf{0}\\
\hline
\end{tabular}
\caption{Results of each model without applying any fairness constraint.}
\label{table:tabel_results_wo_fair_Law_Maths}
\end{table}
\begin{table}[h!]
\begin{tabular}{ |p{1.2cm}|p{1.0cm}|p{1.5cm}|p{1.4cm}|p{1.4cm}|}
\hline
Model &  F1  &  Accuracy & SPD &   EOD\\
\hline
\hline
LR &  \textbf{0.94643}  & \textbf{0.89983} &   -0.100764  &  -0.043942\\\hline
FT&  0.94305   &  0.89230 & 0.001328 & 0.000617 \\\hline
Tab& 0.94371 & 0.89342 &  \textbf{0}   & \textbf{0}  \\
\hline
Perceiver & 0.94237 & 0.89102 & \textbf{0}  & \textbf{0}\\
\hline
SAINT   & 0.94012  & 0.88701 & \textbf{0}   & \textbf{0} \\
\hline
\end{tabular}
\caption{Results of each model with fairness constraints on Law School. Protected attribute: race.}
\label{table:tabel_results_w_fair_Law}
\end{table}
Table \ref{table:tabel_results_wo_fair_Law_Maths} shows that all the transformer-based models (without fairness constraint) have comparable performance on the Law School dataset. Nonetheless, the Tab model (without fairness constraint) has shown marginally better performance scores on the Law School dataset than the other transformer models, with  accuracy of $ 0.90016 $ and F1 score of $0.94664$. Moreover, Table \ref{table:tabel_results_wo_fair_Law_Maths} indicates that the LR model has better performance than the Tab model by a very small margin, with an accuracy improvement by $ + 0.00705$ and F1-score improvement by $+ 0.0032$. Furthermore, Table \ref{table:tabel_results_wo_fair_Law_Maths} and Figure \ref{fig:ABROC_models} show that the SAINT model outperforms the other models in terms of fairness without requiring any explicit debiasing method. This ensures that the candidates are treated equally based on different races (protected attribute in the Law School dataset) when predicting whether or not the candidate will pass the bar exam. In short, the SAINT model emerges as an ideal candidate for correctly identifying successful students at equal rates for different race subgroups.\newline 
However, transformer-based models are nonlinear with a large number of model parameters. This requires a large enough dataset to train \cite{hoffmann2022training} for achieving a good performance. Thus, the number of samples of Student-Mathematics dataset (only 395 samples) is not enough for training these transformer neural networks. As a result, the LR model outperforms the SAINT model on the Student-Mathematics dataset, with an accuracy of $0.93277$ and F1-score of $0.91111$. Nevertheless, same as in the Law School dataset, the SAINT model outperforms the LR model on the Student-Mathematics dataset based on the fairness metrics (SPD, EOD) without requiring any explicit debiasing technique. \newline 
Using the mentioned bias mitigation method in Section \ref{fairness_constraints}, Table \ref{table:tabel_results_w_fair_Law} and Figure \ref{fig:ABROC_models} show that the transformer-based models (with fairness constraint) successfully limit the bias in the final predictions on the Law School dataset at a minimal cost in terms of performance. Intriguingly, the Tab model (with fairness constraint) shows better performance and fairness than the other transformer models, with an accuracy of 0.89342, F1-score of 0.94371, and perfect group fairness. Furthermore, although the SAINT model showed its usefulness in achieving perfect group fairness without requiring any explicit debiasing method, the Tab model (with fairness constraint) outperformed the SAINT model (without fairness constraint) in terms of fairness and performance simultaneously, with an accuracy improvement by $ + 0.00128$ and F1-score improvement by $+ 0.00072$.\newline 
Additionally, the LR model showed two drawbacks: (i) it hardly shows a noticeable better performance than the other transformer models on the Law School dataset, and (ii) Figure \ref{fig:ABROC_models} indicates that the LR model tends to fail to converge to a complete fair solution, with ABROCA of 0.171 for the LR model (without fairness constraint) and ABROCA of 0.0889 for the LR model (with fairness constraint). \newline
In general, previous results in Table \ref{table:tabel_results_wo_fair_Law_Maths} and Table \ref{table:tabel_results_w_fair_Law} show that when we use a large enough dataset, such as the Law School dataset, there is a slight trade-off between the performance and fairness using the transformer models (with fairness constraint). Additionally, Table \ref{table:tabel_results_wo_fair_Law_Maths} indicates that the SAINT model approximates the ideal distribution of the given dataset, which has a negligible accuracy-fairness trade-off.

\section{Conclusion}
Improving fairness while training the model and preserving its sound performance is challenging in ML. The prior works have focused on enhancing fairness in the final predictions using different bias mitigation techniques in classical ML models. Still, as we show, transformer models have some advantages when we use them to study fairness over the tabular dataset. In particular, we showed that the SAINT model achieves perfect group fairness without requiring any explicit debiasing method. Additionally, we showed that the Tab model (with fairness constraint) improves fairness for the protected group in the Law School dataset at a negligible cost in terms of accuracy compared to the other models. Our critical insight is that when trying to improve the fairness in final results using transformer models, it is valuable to check if the tabular dataset is large enough and compatible with the number of the parameters in the transformer model.\\
Consequently, we believe that this insight and use of transformer models for fairness over tabular datasets provide a foundation for pursuing the fairness of artificial intelligence in educational and other domains.\\
For future work, we will comprehensively study the empirical performance of the transformer models for fair classification in different tabular datasets. In addition, we will employ the ideas of transfer learning and self-supervised learning to solve the problem of training the transformer-based models on a small dataset.

\hspace{2.2cm} ACKNOWLEDGMENTS \\
This work has received funding by the European Social Fund via IT Academy programme.

\bibliographystyle{abbrv}
\bibliography{main}

\balancecolumns
\end{document}